# HSI BASED COLOUR IMAGE EQUALIZATION USING ITERATIVE $n^{th}$ ROOT AND $n^{th}$ POWER

Gholamreza Anbarjafari

*iCV Group, IMS Lab, Institute of Technology, University of Tartu, Tartu 50411, Estonia*

*sjafari@ut.ee*

**ABSTRACTS**

In this paper an equalization technique for colour images is introduced. The method is based on $n^{th}$ root and $n^{th}$ power equalization approach but with optimization of the mean of the image in different colour channels such as RGB and HSI. The performance of the proposed method has been measured by the means of peak signal to noise ratio. The proposed algorithm has been compared with conventional histogram equalization and the visual and quantitative experimental results are showing that the proposed method over perform the histogram equalization.

**KEYWORDS-** Image equalization, iterative $n^{th}$ root and $n^{th}$ power equalization, peak signal to noise ratio.

## I. INTRODUCTION

Contrast is the difference in visual properties that makes an object (or its representation in an image) distinguishable from other objects and the background. In visual perception of the real world, contrast is determined by the difference in the colour and brightness of the object with other objects in the same field of view. The human visual system is more sensitive to contrast than absolute luminance; hence, we can perceive the world similarly regardless of the considerable changes in illumination conditions.

If an image is overall very dark or a very bright, the information may be lost in those areas which are excessively and uniformly dark or bright. The problem is how the contrast of an image can be optimized to represent all the information in the input image. Techniques such as histogram equalization, gamma correction, and linear contrast correction have been used to reduce these effects [1- 4].

In many image-processing applications, the standard greyscale histogram equalization (GHE) method is one of the simplest and most effective primitives for contrast enhancement [5], which attempts to produce an output histogram that is uniform [6]. There are also some other techniques which are used for equalizing greyscale images [7-9] such as dynamic histogram equalization (DHE) and singular value equalization (SVE).

One of the disadvantages of the GHE is that the information laid on probability distribution function (PDF) of the image will be totally lost. In [10, 11], it has been shown that the histogram of face images can be used for face recognition. Hence, it is necessary that if any equalization is used, the pattern of the histogram should not be changed dramatically.

Nowadays, colour images are more of interests, and in recent image processing research areas colour images have been used quite frequently. Colour images have introduced much more questions than grey scale ones. Simply the problem will be triple because the image is not being described in a 3D colour space, instead of 1D (grey) colour space. The approaches which are used for equalization of a grey scale image do not work properly for colour images. For instant, Fig. 1 (a) illustrates a low contrast face image from the CALTECH face database [12] and resultant image in (b) after being equalized by using histogram equalization in R, G, and B channel separately and then using them to reconstruct a new equalized image.

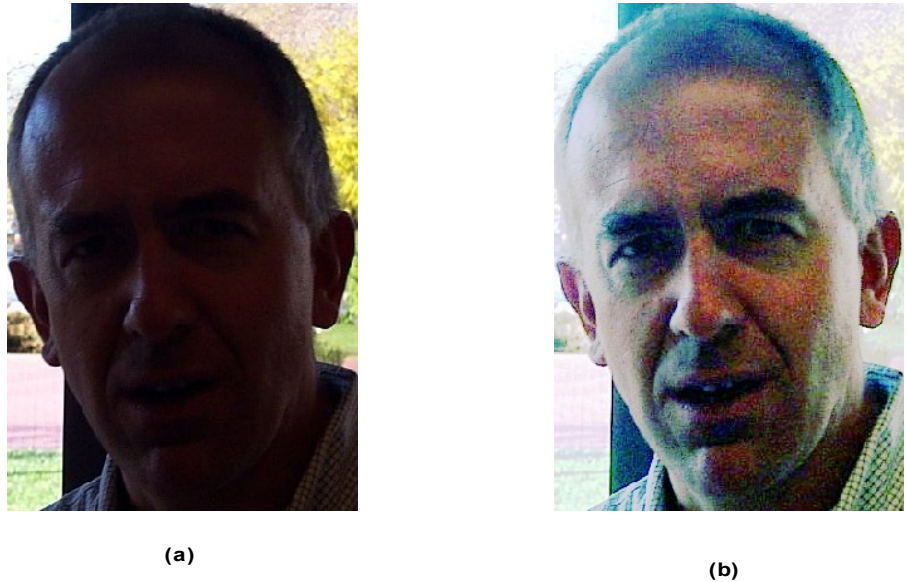

(a) (b)

**Fig. 1**: A face image from the CALTECH face database (a), and the equalized face image using histogram equalization in each R, G, and B channels separately (b).

From Fig. 1 (b) it is very clear that the information laid on the colour of the image such as skin colour, the background details such as leaves of the tree in the right top of the image have been lost. Not only that but also the image, visually has become unreliable due to artificial generated colours.

In this work, we have proposed a new method of image equalization based on iterative equalization of the image by using $n^{th}$ root and $n^{th}$ power. The proposed reduces loss of information laid on the colour pixels and keeps most of the statistics of the colour image. Furthermore, the proposed method has been compared with some other equalization method and has been shown that the proposed technique produces a visually clearer image with small loss of information. The experiment has been tested on some images taken from the internet [13, 14].

## II   METHODS OF EQUALIZING A COLOUR IMAGE

There exist several methods in literature to equalize a colour image [5, 14-19]. The equalization procedure can been done in RGB or HSI colour spaces. Each image can be described by three sub-images (red, green, and blue or hue, saturation, and intensity) in different colour spaces. The general procedure of equalizing an image in RGB colour space is as follow:

1. Decompose the colour image into its R, G, and B sub-images,

2. Equalize each sub-image,

3. Combine all the three new sub-images to attain the new image.

The equalization procedure of a colour image is HSI colour space is similar to the equalization in RGB domain with slight changes. These steps are:

1. Convert the colour image into HSI colour space and decompose it into its hue, saturation and intensity sub-images.

2. Equalize the intensity sub-image.

3. Combine the sub-images to attain a new equalized image.

From the definition of saturation and hue sub-images it is straight forward that equalization of hue or/and saturation sub-images will lead to loss of huge amount of colour information. So the equalization in HSI colour space will be done only by equalizing the intensity sub-image.

# III ITERATIVE n<sup>th</sup>-ROOT AND n<sup>th</sup>-POWER EQUALIZATION

The iterative proposed method for equalization is based on achieving a sub-image with mean of 0.5. First of all R, G, and B or I sub-images of RGB colour space or HSI colour space have been changed from [0,255] range into [0,1] respectively. Then the mean of each sub-image has been calculated. The aim is to change the contrast of the image in the way that the mean of the image with low contrast and low or high intensity changes into 0.5. If the mean is not already 0.5, then the following ratio will be calculated:

$$\Theta = \frac{\ln 0.5}{\ln \mu_\chi} \qquad (1)$$

where $\Theta$ is used as a power factor to find the new image, and $\mu_\chi$ is the mean of image $\chi$. The mean of the image can be calculated as follow:

$$\mu = \frac{\sum_{i=1}^{M}\sum_{j=1}^{N} a_{ij}}{N \times M} \qquad \left(A = [a_{ij}]_{M \times N} \qquad A = \{R, G, B\}\right) \qquad (2)$$

Where $A_{M \times N}$ represents one of the sub-images in the RGB colour space or the intensity sub-image in HSI colour space and $a_{ij}$ is the intensity value of the pixel of the i<sup>th</sup> row and j<sup>th</sup> column of the image. It is obvious that $\Theta$ is less than 1 if $\mu_\chi < 0.5$ and is less than 1 if $\mu_\chi > 0.5$. Fig. 2 shows the value of $\Theta$ respect to $\mu_\chi$.

A new image can be obtained by the following equation:

$$\Xi_{new} = \Xi^\Theta \qquad (3)$$

Where $\Xi_{new}$ and $\Xi$ represent the new image and the old image respectively. If $\Theta < 1$ then $\mu_{\Xi_{new}} < \mu_\Xi$, where $\mu_\Xi \in [0,1]$ so the equation (3) is representing the $\Theta^{th}$ root of the image.

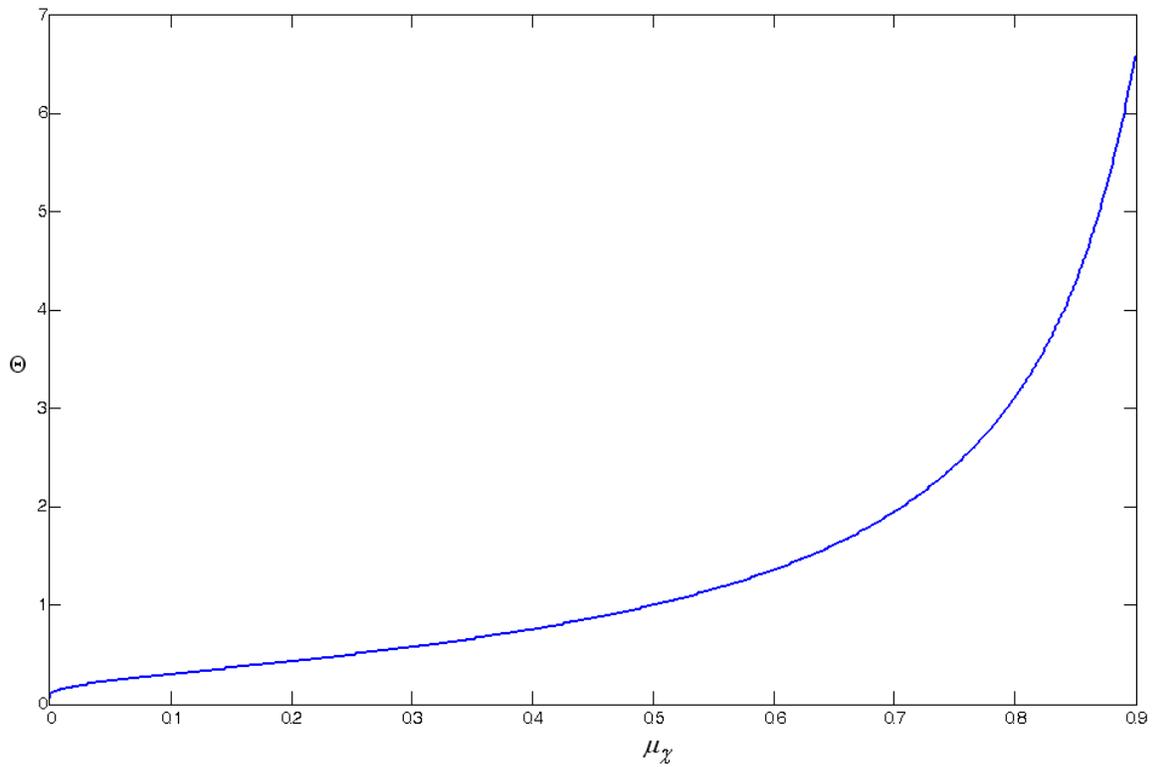

**Fig. 2:** the value of $\Theta$ respect to $\mu_\chi$.

If $\Theta > 1$ then $\mu_{\Xi_{new}} > \mu_\Xi$, so the equation (3) is representing the $\Theta^{th}$ power of the image. Repeating this procedure will be terminated when $\mu_{\Xi_{new}} = 0.5$.

Fig. 3 shows a block diagram of the steps that should be taken to achieve an equalized colour image in RGB or HSI colour spaces. In the next section some experimental results of the proposed technique has been demonstrated on some low contrast colour images.

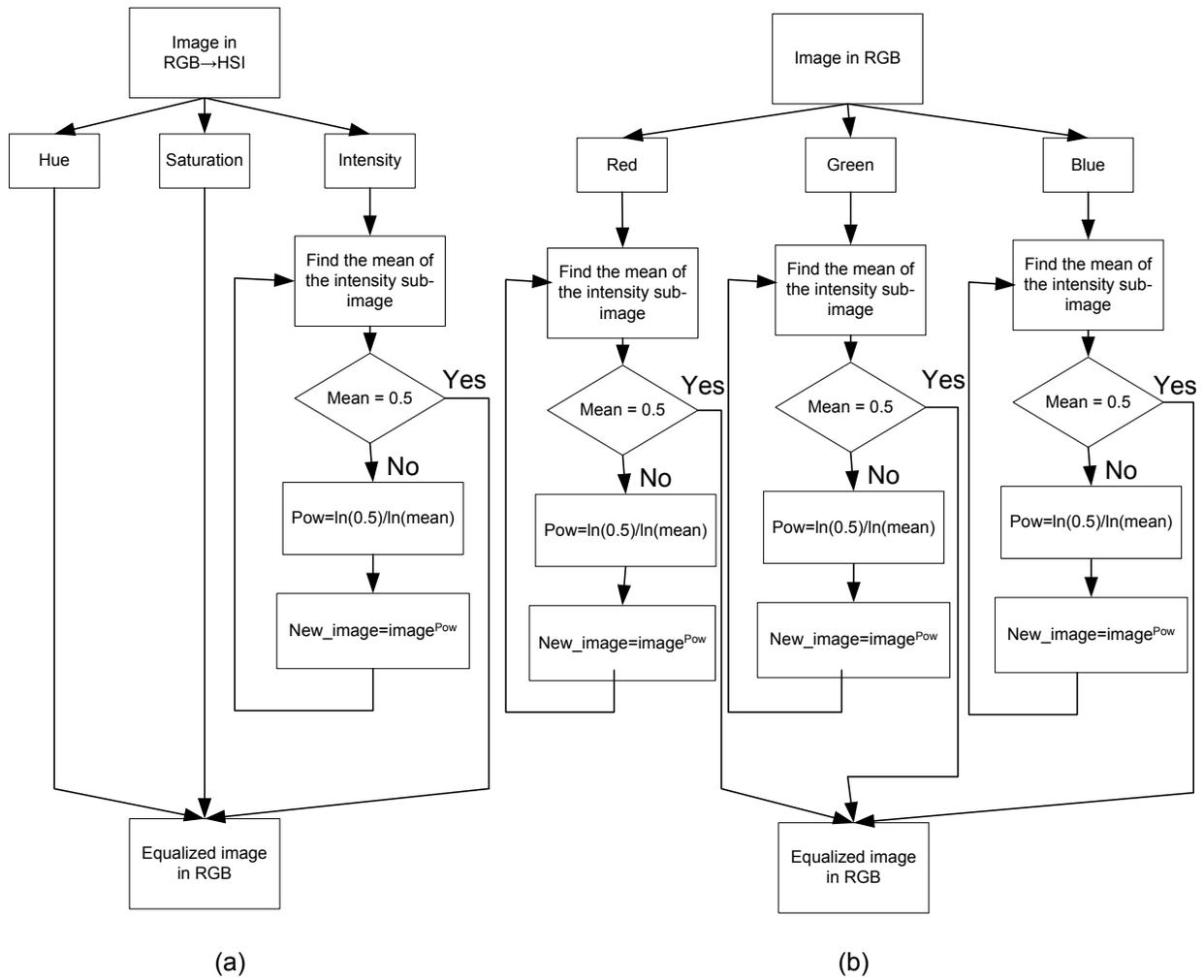

**Fig. 3:** The block diagram of different steps that should be taken to equalize a colour image in HSI (a) and RGB (b) colour spaces.

## IV EXPERIMENTAL RESULTS AND DISCUSSIONS

Fig. 4 illustrates an image with a good contrast and the same scene with low contrast taken from [13]. The background details and also the details around the eyes of the girl are not visible in the low contrast image.

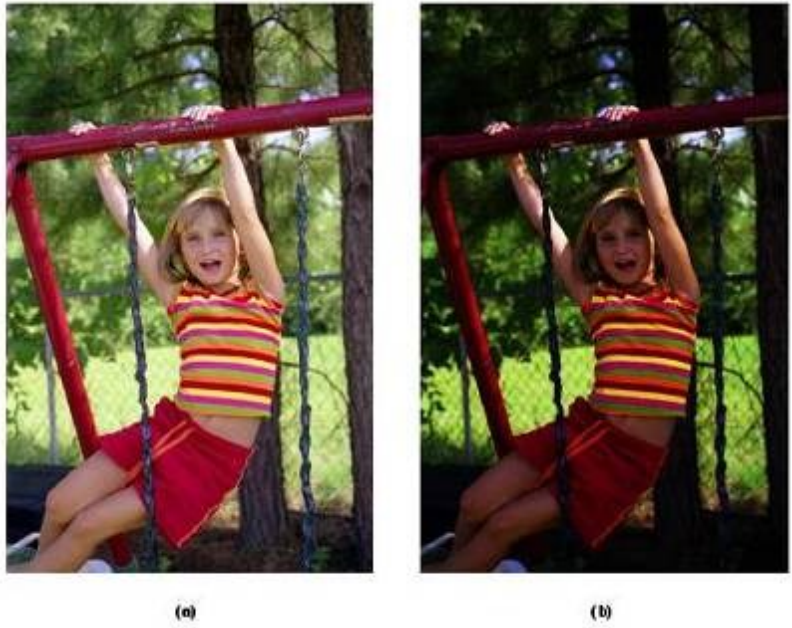

**Fig. 4**: An image taken from [13] with good contrast (a) and low contrast (b).

In order to be able to compare the performance of the iterative $n^{th}$ root and $n^{th}$ power approach over the histogram equalization, the low contrast image in Fig. 4 (b) has been equalized by several combinations equalization techniques such as: the iterative $n^{th}$ root and $n^{th}$ power algorithm in RGB colour space and also in HSI colour space. Also the equalization has been achieved by using histogram equalization in RGB and HSI colour spaces. Fig. 5 shows the equalized image of the low contrast image shown in Fig. 4 (b) by using different methods.

The processes have been repeated for another low contrast image taken from [13] shown in fig. 5. Peak Signal to Noise Ratio (PSNR) has been used to mathematically compare the proposed method with histogram equalization. PSNR can be calculated by using the following equation:

$$PSNR = 20 \times \log_{10}\left(\frac{255}{\sqrt{\varepsilon}}\right)$$

$$\varepsilon = \frac{1}{M \times N} \sum_{i=1}^{M} \sum_{j=1}^{N} \left[\Xi_{original}(i,j) - \Xi_{equalized}(i,j)\right]^2$$

(4)

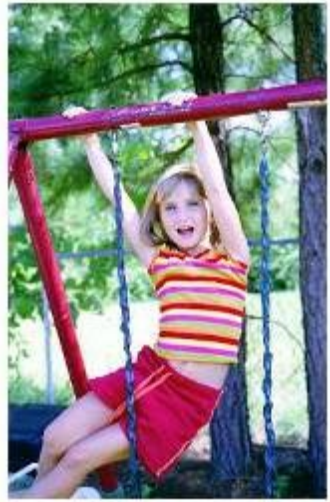 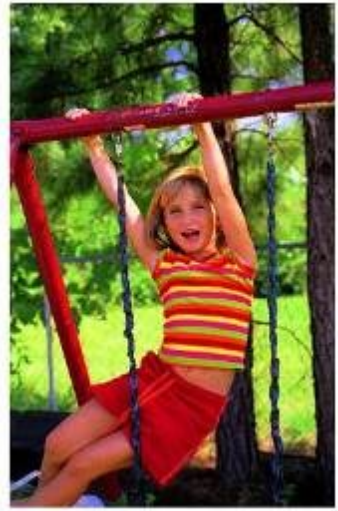
(a) (b)

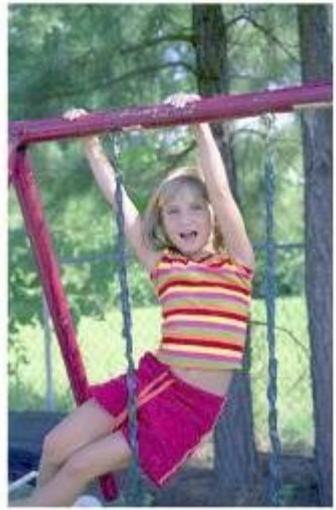 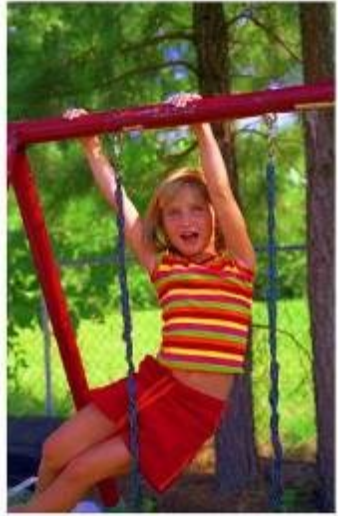
(c) (d)

**Fig. 4:** An equalized image by using histogram equalization in RGB (a), and in HSI (b) colour spaces and by using the proposed method in RGB (c), and in HSI (d) colour spaces.

Where $\epsilon$ is the mean square error between the original image and the equalized image and $M$ and $N$ represent the size of the image. The PSNR values are shown in table 1 for both image shown in Fig. 3 and Fig. 5.

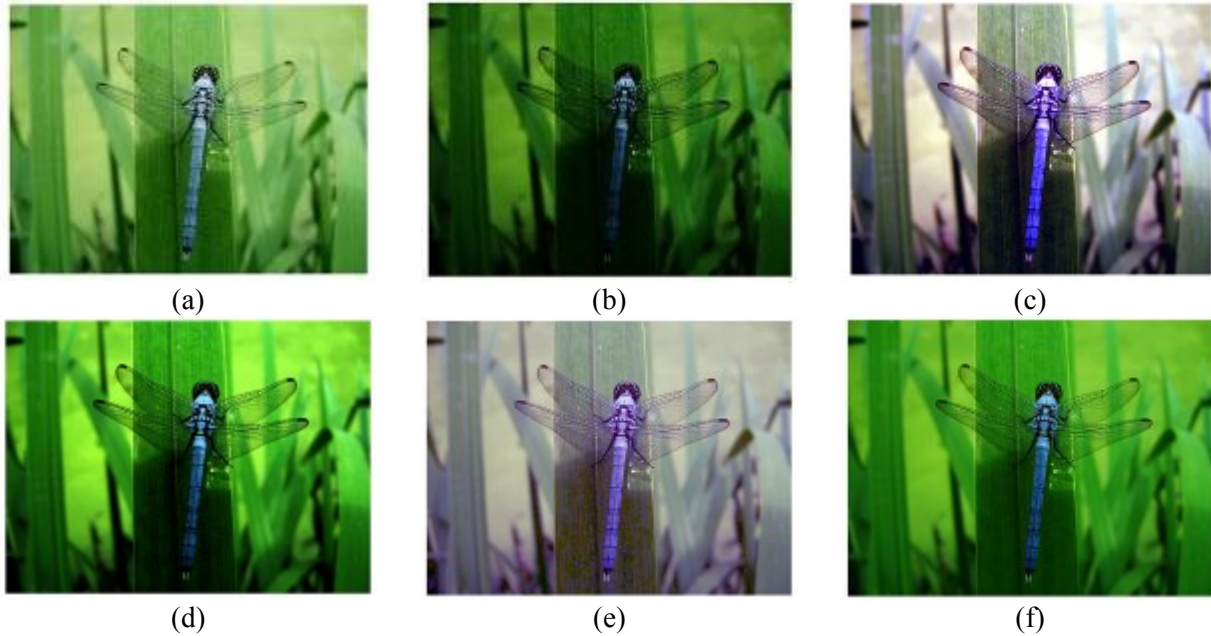

**Fig. 5:** An image taken from [13] (a) and the same image with low contrast (b), the equalized image by using histogram equalization in RGB (c), and in HSI (d) colour spaces and by using the proposed method in RGB (e), and in HSI (f) colour spaces.

The results demonstrate that the proposed method always over perform the histogram equalization. The results show up to 1.98% and 3.63% increase of PSNR value for [13] and [14] respectively.

To demonstrate how the PSNR values are changing in each iteration, the PSNR values of each iterations for equalizing the [13] in RGB colour space has been calculated and is shown in fig. 6. Also fig. 6 shows that the mean of each sub-images is reaching to 0.5.

**Table 1**: comparison between the proposed method and histogram equalization for two different low contrast images [13], [14]

| PSNR | Techniques | | | | | | | |
|---|---|---|---|---|---|---|---|---|
| | The girl's image [12] | | | | The flying dragon image [13] | | | |
| | Histogram Equalization in HSI | Proposed method in HSI | Proposed method in RGB | Histogram Equalization in RGB | Histogram Equalization in HSI | Proposed method in HSI | Proposed method in RGB | Histogram Equalization in RGB |
| R channel | 71.2841 | 72.5282 | 71.5843 | 69.0292 | 62.3284 | 61.6900 | 67.5193 | 65.0037 |
| G channel | 70.2682 | 70.7067 | 65.1439 | 64.8588 | 63.3482 | 67.3805 | 67.6852 | 64.0286 |
| B channel | 66.9479 | 66.0986 | 60.8000 | 59.7304 | 61.3169 | 61.7601 | 60.8322 | 59.8976 |
| Average | 69.5001 | 69.7779 | 65.8427 | 64.5395 | 62.7968 | 63.6102 | 65.3455 | 62.9766 |

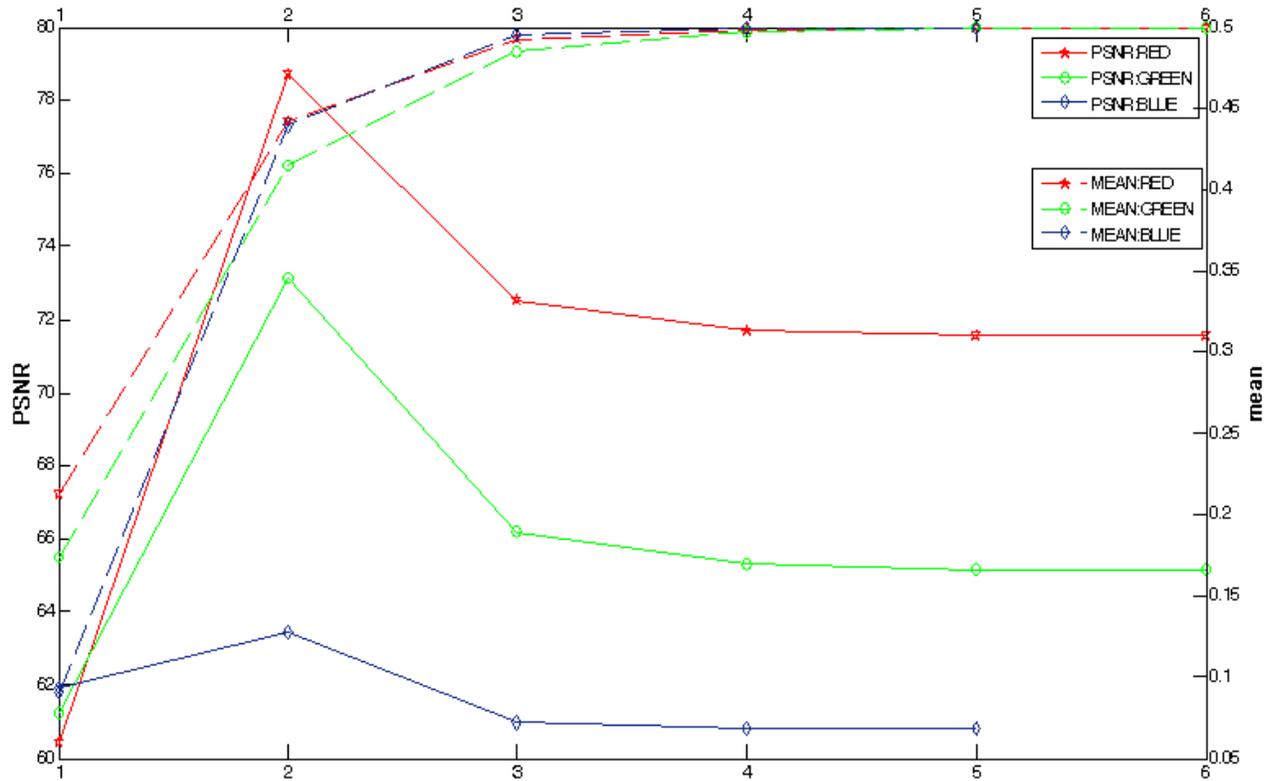

**Fig 6:** The PSNR and mean changes in each iteration of equalizing [13] in RGB domain.

## V  CONCLUSION

In this work an equalization technique for colour images was introduced. The proposed technique was based on nth root and nth power equalization approach to obtain an equalize image in different colour channels such as *RGB* and *HSI*. The performance of the proposed method was measured by calculating PSNR values. The proposed algorithm has been compared with other classical histogram equalization.

# VI   REFERENCES

[1]. W.G. Shadeed, D.I. Abu-Al-Nadi, and M.J. Mismar, Road Traffic Sign Detection in Color Images, *Proc. 10$^{th}$ International Conference on Environmental and Computer Science*, Vol. 2, 2003, pp. 890-893.

[2]. H. Demirel, C. Ozcinar, and G. Anbarjafari, "Satellite image contrast enhancement using discrete wavelet transform and singular value decomposition", *IEEE Geoscience and Remote Sensing Letters*, Vol. 7, 2010, pp. 333-337.

[3]. T. K. Kim, J. K. Paik, and B. S. Kang, "Contrast enhancement system using spatially adaptive histogram equalization with temporal filtering," *IEEE Transactions on Consumer Electronics*, Vol. 44, No. 1, 1998, pp. 82–86.

[4]. H. Demirel, G. Anbarjafari, and M. N. S. Jahromi, "Image equalization based on singular value decomposition", *23$^{rd}$ IEEE International Symposium on Computer and Information Sciences*, 2008, pp. 1-5.

[5]. T. Kim, H.S. Yang, "A Multidimensional Histogram Equalization by Fitting an Isotropic Gaussian Mixture to a Uniform Distribution", *IEEE International Conference on Image Processing*, 8-11 Oct. 2006, pp. 2865 – 2868.

[6]. A. R. Weeks, L. J. Sartor, and H. R. Myler, 'Histogram specification of 24-bit color images in the color difference (C-Y) color space', *Proc. SPIE*, 1999, 3646, pp. 319–329

[7]. M. Abdullah-Al-Wadud, H. Kabir, M. A. A.Dewan, C. Oksam, "A Dynamic Histogram Equalization for Image Contrast Enhancement", *IEEE International Conference on Consumer Electronics (ICCE)*, 10-14 Jan. 2007, pp. 1 – 2.

[8]. C. C. Sun, S. J. Ruan, M. C. Shie, and T, W. Pai, "Dynamic Contrast Enhancement based on Histogram Specification", *IEEE Transactions on Consumer Electronics*, Vol. 51, No. 4, 2005, pp.1300–1305.

[9]. G. Anbarjafari, S. Izadpanahi, C. Ozcinar, and H. Demirel, "Illumination compensation by using singular value decomposition and discrete wavelet transform", *19th IEEE Conference on Signal Processing and Communications Applications*, 2011, pp. 904 - 907.

[10]. H. Demirel, and G. Anbarjafari, "Pose invariant face recognition using probability distribution functions in different color channels", *IEEE Signal Processing Letters*, Vol. 15, 2008, pp. 537-540.

[11]. G. Anbarjafari, "Face recognition using color local binary pattern from mutually independent color channels", *EURASIP Journal on Image and Video Processing*, Vol. 1, 2013, pp. 1-11.

[12]. Frontal face dataset. Collected by Markus Weber at California Institute of Technology, Retrieved Jan. 2011, http://www.vision.caltech.edu/html-files/archive.html

[13]. An image taken from the internet, Retrieved on 10th January 2011 from the World Wide Web: http://www.rangefinderforum.com/forums/showthread.php?t=12318.

[14]. An image taken from the internet, Retrieved on 10th January 2011 from the World Wide Web: http://www.digitalcamerareview.com/assets/4391.

[15]. M. Abdullah-Al-Wadud, M. H. Kabir, A.Dewan, O. Chae, "A Dynamic Histogram Equalization for Image Contrast Enhancement", *IEEE Transactions on Consumer Electronics*, Vol. 53, Issue 2, May 2007, pp. 593 - 600.

[16]. L. Lucchese, and S. K. Mitra, "A new method for color image equalization", *International Conference on Proceedings Image Processing*, 2001, Vol. 1, 7-10 Oct. 2001, pp: 133 – 136.

[17]. A. K. Forrest, "Colour histogram equalisation of multichannel images", *IEE Proc. Vision, Image and Signal Processing*, Vol. 152, Issue 6, 9 Dec. 2005, pp: 677 – 686.

[18]. C. Ozcinar, H. Demirel, and G. Anbarjafari, "Image Equalization Using Singular Value Decomposition and Discrete Wavelet Transform", *Discrete Wavelet Transforms: Theory and Applications*, 2011, pp. 87-94.

[19]. T. Kim, and H. S. Yang, "Color histogram equalization via least-squares fitting of isotropic Gaussian mixture to uniform distribution", *Electronics Letters*, 2006, 42(8), pp. 452-453.